%% file: neurips_2026.tex
\documentclass{article}

\PassOptionsToPackage{numbers,compress,sort}{natbib}

\usepackage[preprint]{neurips_2026}

\input{preamble}
\newcommand{\method}{HyperSafe}
\newcommand{\ssn}{SSN}

\title{HyperSafe: Inference-Time Safety Recovery for Fine-Tuned Language Models}

\hypersetup{
  pdftitle={HyperSafe: Inference-Time Safety Recovery for Fine-Tuned Language Models},
  pdfauthor={Aznaur Aliev, Carlos Hinojosa, Abdelrahman Eldesokey, Bang An, Bernard Ghanem, Yibo Yang}
}

\author{%
  \textbf{Aznaur Aliev \quad Carlos Hinojosa \quad Abdelrahman Eldesokey} \\
  \textbf{Bang An \quad Bernard Ghanem \quad Yibo Yang} \\
  King Abdullah University of Science and Technology, Saudi Arabia \\
  {\small\texttt{\{aznaur.aliev,carlos.hinojosa,yibo.yang\}@kaust.edu.sa}}
}

\begin{document}

\raggedbottom
\setlength{\floatsep}{6pt plus 1pt minus 1pt}
\setlength{\textfloatsep}{12pt plus 2pt minus 2pt}

\maketitle

\input{sec/0_abstract}
\input{sec/1_introduction}
\input{sec/2_related_works}
\input{sec/3_method}
\input{sec/4_experiments}
\input{sec/5_conclusions}

\renewcommand{\acksection}{\section*{Acknowledgments}}
\begin{ack}
This research was supported by King Abdullah University of Science and
Technology (KAUST), Center of Excellence for Generative AI, under Award
No.~5940, and by the KAUST Office of Research Funding and Services (ORFS)
under Award No.~ORFS-CRG13-2025-6903.
\end{ack}

\bibliographystyle{unsrtnat}    
\bibliography{custom}

%
%
%
%
%
%
%
%
%
%


\clearpage
\input{sec/6_appendix}

\end{document}

%% file: preamble.tex
\usepackage{times}
\usepackage{latexsym}

\usepackage[T1]{fontenc}

\usepackage[utf8]{inputenc}

\usepackage{xurl}
\usepackage[
    colorlinks=true,
    linkcolor=blue,
    citecolor=blue,
    urlcolor=blue
]{hyperref}

\usepackage{microtype} 

\usepackage{inconsolata}

\usepackage{graphicx}

\usepackage{amsmath}
\usepackage{amssymb}
\usepackage{amsfonts}
\usepackage{nicefrac}       
\usepackage{booktabs}
\usepackage{multirow}
\usepackage{enumitem}      
\usepackage{xcolor}
\usepackage{colortbl}
\usepackage{subcaption}
\usepackage{float}

\definecolor{goodgreen}{HTML}{E8F5E9}
\definecolor{badred}{HTML}{FFEBEE}
\definecolor{mutegray}{HTML}{F5F5F5}
\definecolor{warnred}{HTML}{C62828}    
\definecolor{wingreen}{HTML}{2E7D32}   

\newcommand{\mysection}[1]{\vspace{2pt}\noindent\textbf{#1}}

%% file: sec/0_abstract.tex

\begin{abstract}
Safety alignment in large language models can be fragile under fine-tuning, as even benign task adaptation may increase harmful compliance. Existing defenses mainly follow two directions: they either intervene during or after fine-tuning through retraining or weight modification, which can be costly and may hurt task performance, or they use model-agnostic safety classifiers, which may miss failures specific to a given fine-tuned checkpoint. These limitations motivate a post-hoc, model-specific, and non-invasive approach to safety restoration. To meet these requirements, we propose HyperSafe, a framework that restores safety behavior by generating a model-specific Safe Side Network (SSN) for each fine-tuned checkpoint. HyperSafe uses layer-wise activation fingerprints to capture how fine-tuning changes the model's inner representations. With a small set of given calibration prompts, the hypernetwork maps these fingerprints to the parameters of the \ssn{} in a single forward pass. The generated \ssn{} runs alongside the frozen fine-tuned model and performs prompt-level safety classification: harmful prompts are routed to refusal, while safe prompts are answered by the original fine-tuned model. Thus, HyperSafe requires no gradient updates, no safety data at deployment time, and no modification to the deployed model weights. We evaluate HyperSafe on two model families, Qwen2-7B and LLaMA-3-8B, across multiple safety benchmarks. HyperSafe reduces harmful response rates from 19--31\% to below 1\% on every held-out checkpoint, while keeping downstream task accuracy within 1\% of the fine-tuned baseline on average. 
Code is available at \url{https://github.com/nokronim/project-safety-remedy}
\end{abstract}

%% file: sec/1_introduction.tex
\section{Introduction}
\label{sec:intro}

Fine-tuning safety-aligned LLMs is a common way to adapt them to downstream tasks, but it can also weaken their safety alignment. Prior work has shown that, even when the fine-tuning data is totally benign, the harmful-response rate of a LoRA fine-tuned model can climb from under 5\% (the aligned baseline) to over 40\% on standard safety benchmarks~\cite{qi2024finetuning}. One possible reason is that safety alignment can depend heavily on the model's behavior in the first few output tokens~\cite{qi2025safety}. As a result, even benign fine-tuning updates may change the model's refusal behavior and reduce its ability to reject harmful requests.

Existing defenses for this alignment degradation can be grouped into three main categories. The first category applies safety-preserving methods during fine-tuning. These methods add safety constraints during task adaptation, for example, through regularization, safety data, or constrained optimization (Lisa~\cite{huang2024lisa}, RepNoise~\cite{rosati2024representation}, Constrained-SFT~\cite{qi2025safety}). They aim to preserve the base model's refusal behavior while learning the downstream task. However, they often introduce a trade-off between safety and utility: stronger safety constraints may reduce task performance, while weaker constraints may still allow harmful responses. They also require control over the fine-tuning process, so they cannot be easily applied to fine-tuned checkpoints that have already been released.

The second category modifies the model weights after fine-tuning. These methods aim to restore safety by editing, projecting, or merging fine-tuned weights (Safe LoRA~\cite{hsu2024safe}, OneShot SFT~\cite{zhang2026safety}, Task Arithmetic~\cite{ilharco2023task}). They are useful when the model has already been fine-tuned, but they directly change the deployed weights. As a result, their modification may cause forgetting in downstream task performance. They may also require access to the original aligned base model at deployment time, and they still face the same basic safety--utility trade-off as fine-tuning-stage methods.

The third category uses model-agnostic external classifiers. These methods place a separate safety model before or after the LLM to filter harmful inputs or outputs (Llama Guard~\cite{inan2023llama}, WildGuard~\cite{han2024wildguard}, Granite Guardian~\cite{padhi2024granite}). They are easy to deploy because they do not modify the target model or require access to its fine-tuning process. However, since they are not tailored to a specific fine-tuned model, they may miss failures caused by the particular fine-tuning recipe, such as harmful responses triggered by newly learned task-specific syntax or domain patterns. They may also spend capacity detecting risks that are not relevant to the deployed model.

Overall, existing defenses either require intervention during or after fine-tuning, which can be costly and may reduce task utility, or rely on model-agnostic filters that may not capture failures specific to a given fine-tuned model.


Based on these limitations, we ask the following question:
\begin{quote}
\textit{Can we recover the safety behavior of the original aligned model for a given fine-tuned checkpoint using a model-specific lightweight post-hoc method, without retraining, modifying the deployed weights, or degrading downstream task performance?}
\end{quote}

This question leads to three design requirements. First, it should not require access to or control over the original fine-tuning process. Second, it should not modify the deployed fine-tuned weights, since doing so may damage the task behavior learned during fine-tuning. Third, it should be specific to the fine-tuned checkpoint, rather than relying on a fixed external classifier that ignores how the model was adapted. These requirements motivate the use of a model-conditioned side network. Instead of changing the fine-tuned model itself, we attach a lightweight \emph{Safe Side Network} (\ssn{}) that runs alongside the frozen model and decides whether to answer or refuse a prompt. Since the \ssn{} is separate from the task model, the fine-tuned model can preserve its downstream ability, while the \ssn{} focuses on recovering safety behavior. To avoid training a new \ssn{} for every released checkpoint, we use a hypernetwork to generate the \ssn{} weights from the fine-tuned model's activation fingerprints.

\input{tables/comparison_intro}

\begin{figure*}[t]
\centering
\includegraphics[width=\textwidth]{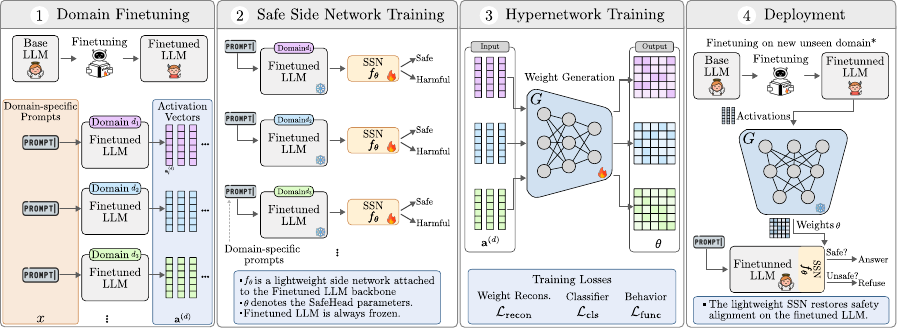}
\caption{\method{} pipeline. \textbf{Stages 1--3 (training, one-time):} fine-tune the base model on each training domain (1), train a domain-specific \ssn{} per fine-tuned checkpoint (2), and train the hypernetwork to map activation fingerprints to \ssn{} weights (3). \textbf{Stage 4 (deployment):} for any new fine-tuned model, extract calibration activations and generate a model-specific \ssn{} in a single forward pass.}
\label{fig:pipeline}
\end{figure*}

We introduce \method{}, a post-hoc framework for restoring safety in fine-tuned LLMs without retraining or modifying the deployed model weights. Given a fine-tuned checkpoint, \method{} first extracts layer-wise activation fingerprints from a small set of calibration prompts. A hypernetwork then maps these fingerprints to generate the weights of a lightweight \ssn{}. The \ssn{} is then attached to the frozen fine-tuned model and produces a prompt-level safety score. At inference time, prompts predicted to be harmful are routed to a refusal response, while safe prompts are handled by the original fine-tuned model. In this way, \method{} is checkpoint-specific, because the generated \ssn{} depends on the target model's activations; post-hoc, because it is applied after fine-tuning; and non-invasive, because it does not update or edit the deployed model weights.

\mysection{Contributions.} We summarize our main contributions as follows:
\begin{enumerate}[leftmargin=*,topsep=2pt,itemsep=2pt]
    \item We formulate post-hoc safety restoration as the activation-conditioned generation of \ssn{}. Given a fine-tuned checkpoint, our method generates a model-specific \ssn{} from its activation fingerprints, without additional gradient updates or safety data at deployment time.

    \item We design \method{}, a hypernetwork-based framework that predicts the weights of a lightweight \ssn{}. To make this prediction feasible, \method{} uses layer-wise activation conditioning, direction-magnitude decomposition, and factorized weight generation.

    \item We show that the generated \ssn{} can restore refusal behavior while preserving downstream task performance. Across Qwen2-7B and LLaMA-3-8B, the hypernetwork is trained on fine-tuned models from training domains and evaluated zero-shot on held-out fine-tuned checkpoints. On three safety benchmarks, \method{} reduces harmful response rates from 19--31\% to below 1\% on every held-out checkpoint, while keeping downstream task accuracy within 1\% of the fine-tuned baseline on average.
\end{enumerate}


%% file: tables/comparison_intro.tex
\begin{table}[t]
\caption{Comparison with existing defenses against fine-tuning-induced safety degradation. \emph{Setup cost per checkpoint}: compute to deploy on each new fine-tuned model. \emph{Model-aware}: tailored to the target rather than a one-size-fits-all classifier. \emph{Preserves model weights}: deployed weights kept unchanged. \emph{Standalone at deploy}: requires only the fine-tuned model at inference, no aligned base $\mathcal{M}_0$. \method{} is the only approach that satisfies all four.}
\label{tab:comparison}
\centering
\resizebox{0.8\columnwidth}{!}{%
\begin{tabular}{@{}lcccc@{}}
\toprule
\textbf{Method} & \textbf{Setup cost} & \textbf{Model-} & \textbf{Preserves} & \textbf{Standalone} \\
                & \textbf{per checkpoint} & \textbf{aware}  & \textbf{model weights} & \textbf{at deploy} \\
\midrule
Vaccine~\citep{yi2024vaccine}             & full train     &  \checkmark &  $\times$  &  \checkmark \\
SafeInstr~\cite{bianchi2024safetytuned}  & partial train  &  \checkmark &  $\times$  &  \checkmark \\
Safe LoRA~\cite{hsu2024safe}             & projection     &  \checkmark &  $\times$  &  $\times$    \\
OneShot~\cite{zhang2026safety}           & 1-example SFT  &  \checkmark &  $\times$  &  \checkmark  \\
Llama Guard 4~\cite{meta2025llama4}      & none           &  $\times$   &  \checkmark &  \checkmark  \\
\midrule
\textbf{\method{} (ours)}                & 1 fwd pass     &  \checkmark &  \checkmark &  \checkmark  \\
\bottomrule
\end{tabular}%
}
\end{table}

%% file: sec/2_related_works.tex
\section{Related Work}
\label{sec:related_works}

\paragraph{Safety alignment of LLMs.}
Modern LLMs acquire safety alignment through supervised fine-tuning on curated demonstration data~\cite{ouyang2022training, bai2022training}, reinforcement learning from human feedback~\cite{stiennon2020summarization, christiano2017deep}, Constitutional AI~\cite{bai2022constitutional}, and direct preference optimization~\cite{rafailov2023dpo}, producing aligned models such as LLaMA-3-Instruct~\cite{dubey2024llama} and Qwen2-Instruct~\cite{yang2024qwen2}. However, this alignment is shallow: \citet{qi2025safety} shows that refusal behavior is encoded in only a few token positions, and \citet{wei2024assessing} finds that safety lives in low-rank subspaces that adversarial fine-tuning can easily traverse. \citet{qi2024finetuning} first demonstrated that fine-tuning aligned LLMs, even on benign data, compromises safety, an effect later confirmed across architectures, scales, and PEFT recipes~\cite{fraser2025finetuninglowers, lermen2024lora, zhan2024removing}. Rather than proposing a new alignment recipe, our work focuses on recovering safety after fine-tuning has degraded it.

\paragraph{Defenses against fine-tuning-induced safety degradation.}
Existing defenses fall into three families. {Fine-tuning-stage methods} modify the optimization process during fine-tuning (Lisa~\cite{huang2024lisa}, Vaccine~\cite{yi2024vaccine}, RepNoise~\cite{rosati2024representation}, SafeInstr~\cite{bianchi2024safetytuned}, Constrained-SFT~\cite{qi2025safety}, Booster~\cite{huang2024booster}). \emph{Post-fine-tuning weight-modification methods} edit a deployed checkpoint by projection or arithmetic on its weights (Safe LoRA~\cite{hsu2024safe}, OneShot SFT~\cite{zhang2026safety}, Task Arithmetic~\cite{ilharco2023task, yadav2023ties}, RESTA~\cite{bhardwaj2024resta}, backdoor-enhanced alignment~\cite{wang2024backdooralignment}). \emph{Model-agnostic external classifiers} filter inputs/outputs without inspecting the target model (Llama Guard~\cite{inan2023llama}, ShieldGemma~\cite{zeng2024shieldgemma}, WildGuard~\cite{han2024wildguard}, AEGIS~\cite{ghosh2024aegis}, Granite Guardian~\cite{padhi2024granite}, production moderation~\cite{markov2023moderation, lees2022newgeneration}); a related line monitors a model's own hidden states~\cite{jiang2025hiddendetect, azaria2023internal, arditi2024refusal, zou2023representation}. {Unlike all three families, \method{} generates a model-specific safety classifier at inference time: it requires no per-checkpoint retraining, leaves the deployed weights unchanged, and adapts to recipe-specific shifts that model-agnostic classifiers cannot see.}

\paragraph{Hypernetworks and side networks.}
Hypernetworks~\cite{ha2017hypernetworks, chauhan2024brief} generate the weights of a target network conditioned on context, with applications in continual learning~\cite{vonoswald2020continual}, multi-task adaptation~\cite{navon2023equivariant}, parameter-efficient fine-tuning~\cite{phang2023hypertuning, mahabadi2021parameterefficient}, and zero-shot model construction~\cite{knyazev2021graphhypernet}. Recently, \citet{liang2025dragdrop} scaled the paradigm to LLMs, generating LoRA parameters from text descriptions for zero-shot \emph{utility} adaptation. Ladder Side-Tuning~\cite{sung2022lst} introduces a lightweight side network connected to a frozen backbone via lateral connections for parameter-efficient transfer. Our work generates safety weights from \textit{model activations} (rather than text prompts) and combines hypernetworks with side networks for safety restoration in fine-tuned LLMs; to our knowledge, the first such combination.

%% file: sec/3_method.tex
\section{Method}
\label{sec:method}

Our proposed \method{} framework generates a model-specific \ssn{} for recovering safety in fine-tuned language models. The \ssn{} is a lightweight network attached to the frozen backbone that performs prompt-level safety classification, and its weights are predicted by a hypernetwork from activation fingerprints of the target model. During training, we collect pairs of fine-tuned models and their corresponding \ssn{}s across multiple domains to supervise the hypernetwork. At inference, we produce an \ssn{} for a fine-tuned model on an unseen domain whose safety alignment may have degraded, using our proposed \method{}. In the following sections, we describe the problem setup, the architecture, and the hypernetwork training and deployment procedure.


\subsection{Problem Formulation}

Let $\mathcal{M}_0$ be a safety-aligned base LLM and $\mathcal{M}_d$ a fine-tuned version of $\mathcal{M}_0$ on dataset $d$. Fine-tuning may weaken or alter the safety behavior of $\mathcal{M}_{d}$, leading to compliance with harmful prompts that $\mathcal{M}_0$ would refuse. We seek a generator $G$ that, given any fine-tuned model $\mathcal{M}_{d^*}$ from a potentially unseen dataset, produces an \ssn{} $f_\theta$ parameterized by the generated weights $\theta$:
\begin{equation}
    \theta = G\big(\mathcal{A}(\mathcal{M}_{d^*})\big), \qquad \lambda = f_\theta\big(x \,;\, \mathcal{M}_{d^*}\big),
\end{equation}
where $x$ denotes the input prompt, $\mathcal{A}(\cdot)$ denotes activation fingerprints extracted from a small calibration set (Eq.~\ref{eq:all_activations}), $\theta$ are the \ssn{} parameters predicted by $G$, and $\lambda \in [0,1]$ is a safety score ($\lambda \approx 0$: safe; $\lambda \approx 1$: harmful). The semicolon in $f_\theta(x \,;\, \mathcal{M}_{d^*})$ marks $\mathcal{M}_{d^*}$ as a contextual input: the \ssn{} reads ladder-fused backbone activations from $\mathcal{M}_{d^*}$ when scoring $x$, so $\lambda$ depends on both the prompt and the target fine-tuned model. In our framework, $G$ is implemented as a hypernetwork that predicts $\theta$ from the activation fingerprint of the fine-tuned model (see Section \ref{sec:hypernet}).


\subsection{\method{} Architecture}
\label{sec:arch}

The \ssn{} $f_\theta$ attaches to the frozen backbone as a lightweight branch that predicts a safety score for each prompt and determines whether the model should produce a response or refuse. We build on the Ladder Side-Tuning (LST) architecture~\cite{sung2022lst} and extend it for safety classification. Figure~\ref{fig:architecture} illustrates the overall design.

\input{figures/fig_architecture}

\mysection{Side Transformer.} The \ssn{} is implemented as a small Transformer that processes ladder-fused representations derived from the backbone hidden states. It consists of $K$ layers with hidden dimension $h_s = h/r$ for backbone hidden size $h$ and reduction factor $r$, adding 3--4\% of backbone parameters at the 7--8B scale we evaluate (per-architecture values in Appendix~\ref{app:implementation}, Table~\ref{tab:hyperparams}).

\mysection{Ladder Connections.} Each side layer $k$ fuses downsampled backbone hidden states with side hidden states via a per-layer gate:
\begin{equation}
    \mathbf{h}_{k}^\text{in} = \mu_k \cdot W^\text{down}\, \mathbf{h}_{\phi(k)}^\text{back} + (1{-}\mu_k) \cdot \mathbf{h}_{k-1}^\text{side}
\end{equation}
where $\phi(k)$ maps side layer $k$ to a backbone layer, $\mathbf{h}_{\phi(k)}^\text{back}$ denotes the corresponding backbone hidden state, $\mathbf{h}_{k-1}^\text{side}$ denotes the \ssn{} hidden state from the previous layer, and $\mu_k = \sigma(g_k/\tau)$ is a per-layer gate. The scalar $g_k \in \mathbb{R}$ is one of the parameters generated by the hypernetwork (see Section \ref{sec:hypernet}); $\tau$ is a fixed temperature hyperparameter.

\mysection{Selective Routing.} After the final side layer, hidden states are mean-pooled and passed through an MLP to produce $\lambda$. At inference, prompts scoring above the threshold are routed to a refusal response; those below bypass the \ssn{} entirely, so the base model runs unmodified. This selective routing preserves task accuracy because safe prompts never interact with the \ssn{}.

\subsection{Activation-Conditioned Hypernetwork}
\label{sec:hypernet}

We now describe how the \ssn{} parameters are generated automatically from activation fingerprints. The hypernetwork $G$ takes the activation fingerprint of a fine-tuned model as input and predicts the parameters of the \ssn{}. To make the high-dimensional weight prediction feasible, we employ two complementary design strategies.

\mysection{Activation Fingerprint Extraction. } For each model $\mathcal{M}_d$, we run $N_\text{cal}$ domain-specific calibration prompts and average the backbone hidden state at each layer $l = 1, \dots, L$ after the final token:
\begin{equation}
    \mathbf{a}_l = \frac{1}{N_\text{cal}} \sum_{i=1}^{N_\text{cal}} \mathbf{h}_l(x_i) \in \mathbb{R}^{h}.
\end{equation}
These $L$ vectors form a compact behavioral fingerprint of how fine-tuning shifted the model's representations:
\begin{equation}
    \mathcal{A}(\mathcal{M}_d) = \{\mathbf{a}_l\}_{l=1}^L.
    \label{eq:all_activations}
\end{equation}

\mysection{Direction-Magnitude Decomposition.} To enable more structured conditioning of the hypernetwork, we separate directional changes in representation space from their scale. Specifically, each activation vector is split into a unit direction and log-magnitude:
\begin{equation}
    \hat{\mathbf{d}}_l = \frac{\mathbf{a}_l}{\|\mathbf{a}_l\|}, \quad m_l = \log \|\mathbf{a}_l\|.
\end{equation}
The direction captures the nature of the representational change, while the magnitude reflects its strength. We then compute the vectors:
\begin{align}
\mathbf{z}_\text{dir} &= \text{Attn}\big(\{W_\text{proj}\,\hat{\mathbf{d}}_l\}_{l=1}^L\big),\\
\mathbf{z}_\text{mag} &= \text{MLP}\big(\{m_l\}_{l=1}^L\big),
\end{align}
where $W_\text{proj}$ is a learnable projection matrix. The fused representation $\mathbf{z} = [\mathbf{z}_\text{dir};\, \mathbf{z}_\text{mag}]$ aggregates information across all layers and feeds the weight-generation modules.

\mysection{Layer-wise Factorized Generation.} We implement the hypernetwork $G$ as a global encoder that produces the embedding $\mathbf{z}$ together with a set of layer-wise generators $\{H_k\}_{k=1}^K$ that predict the parameters of each side layer. Rather than predicting all parameters jointly, we generate each side layer's weights independently from the corresponding backbone activation:
\begin{equation}
    \theta_k = H_k(\mathbf{a}_{\phi(k)}, \mathbf{z}),
\end{equation}
where $\phi: \{1,\dots,K\} \rightarrow \{1,\dots,L\}$ maps each side layer $k$ to its corresponding backbone layer. $\mathbf{a}_{\phi(k)}$ provides layer-specific information and $\mathbf{z}$ captures global cross-layer context. Each module $H_k$ is implemented as a lightweight MLP that projects and fuses the local activation $\mathbf{a}_{\phi(k)}$ with the global embedding $\mathbf{z}$ to generate the parameters of side layer $k$. The modules $H_k$ share most parameters across layers but include layer-specific adapters, reducing the effective output dimensionality per prediction.

\mysection{Classifier Generation.} The \ssn{} classifier is generated via a dedicated branch that aggregates information across all layers using cross-layer attention, producing a global embedding of the model's behavioral shift. This embedding is mapped through an MLP with a low-rank output factorization for efficiency (factorization rank in Appendix~\ref{app:implementation}, Table~\ref{tab:hyperparams}).

\subsection{Training and Deployment}
\label{sec:deploy}

Training proceeds in two stages: (1) learning domain-specific \ssn{}s to construct supervision targets, and (2) training the hypernetwork to predict these weights from activation fingerprints.

\mysection{Ground-truth SSN Collection.}\label{sec:gt}
We collect supervised pairs $\{(\mathcal{M}_d, \theta_d^*)\}_{d=1}^D$ by training one \ssn{} per dataset. For each dataset $d$, we first fine-tune the base model $\mathcal{M}_0$ using LoRA~\cite{hu2022lora} and merge the adapter to obtain $\mathcal{M}_d$. We then freeze $\mathcal{M}_d$ and attach a randomly initialized \ssn{}. The \ssn{} is trained on a mixture of per-domain data and category-balanced harmful data from BeaverTails~\cite{ji2024beavertails} (mixture sizes in Appendix~\ref{app:implementation}, Table~\ref{tab:hyperparams}).
Training uses a selective loss: harmful prompts receive both language-modeling loss and classification loss, while safe prompts receive only classification loss. This encourages the \ssn{} to learn when to intervene without degrading task performance on safe inputs. All \ssn{}s share the same initialization, ensuring weight-space alignment for hypernetwork supervision.

\mysection{Hypernetwork Training.} The hypernetwork is trained on $D_\text{train}$ datasets using three complementary losses, weighted by coefficients $\beta_\text{recon}$, $\beta_\text{cls}$, and $\beta_\text{func}$:
\begin{equation}
    \mathcal{L} = \beta_\text{recon}\,\mathcal{L}_\text{recon} \;+\; \beta_\text{cls}\,\mathcal{L}_\text{cls} \;+\; \beta_\text{func}\,\mathcal{L}_\text{func}.
\end{equation}
$\mathcal{L}_\text{recon}$ is the mean-squared error between the generated and ground-truth \ssn{} weights $\hat\theta$ and $\theta^*$, averaged over the generated parameter blocks. $\mathcal{L}_\text{cls}$ and $\mathcal{L}_\text{func}$ are binary cross-entropy losses on the predicted routing score $\hat\lambda(x)$ against the harmful flag $m\in\{0,1\}$, evaluated respectively on the calibration set $\mathcal{C}_d$ used to extract the activation fingerprint and on a held-out probe set $\mathcal{P}_d$ disjoint from $\mathcal{C}_d$; the latter verifies that the generated classifier generalizes beyond the calibration prompts the hypernetwork conditioned on. Hyperparameters and per-architecture training settings are listed in Appendix~\ref{app:implementation}, Table~\ref{tab:hyperparams}.

\mysection{Deployment.} Given a new fine-tuned model $\mathcal{M}_{d^*}$: (1)~extract layer-wise activations from $N_\text{cal}$ calibration prompts; (2)~run one forward pass through the hypernetwork to obtain \ssn{} weights; (3)~load the \ssn{} alongside $\mathcal{M}_{d^*}$. No safety data, gradients, or model modification are required at deployment.

%% file: figures/fig_architecture.tex
\begin{figure*}[t]
\centering
\includegraphics[width=\linewidth]{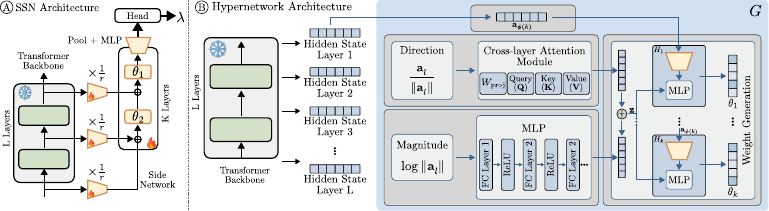}
\caption{\ssn{} architecture. The frozen backbone's hidden states are downsampled via gated ladder connections into a lightweight side network ($\sim$3\% of backbone parameters). The \ssn{} produces a safety score $\lambda$ that routes safe prompts to the base model output and harmful prompts to a refusal response.}
\label{fig:architecture}
\end{figure*}

%% file: sec/4_experiments.tex
\section{Experiments}
\label{sec:experiments}


\mysection{Models and Domains.} We evaluate on Qwen2-7B-Instruct~\cite{yang2024qwen2} and LLaMA-3-8B-Instruct~\cite{dubey2024llama} across 22 datasets spanning QA, commonsense reasoning, math, knowledge, classification, code, and instruction-following (Appendix~\ref{app:domains}). Each dataset is fine-tuned with LoRA (rank 16, $\alpha{=}32$) and merged.

\paragraph{Safety Benchmarks.}
We evaluate on three harmful-prompt benchmarks: \textbf{BeaverTails}~\cite{ji2024beavertails} (700 prompts, 14 categories) is used as the harmful data source during \ssn{} training (see Section~\ref{sec:gt}); the evaluation split is disjoint from the training split. \textbf{AdvBench}~\cite{zou2023universal} (520 prompts), and \textbf{HEx-PHI}~\cite{qi2024finetuning} (330 prompts, 11 categories) are never seen in any training phase, serving as zero-shot safety tests.

\paragraph{Evaluation Protocol.}
We use three holdout splits (per-split domain composition in Appendix~\ref{app:splits}, Table~\ref{tab:splits}; the split $\rightarrow$ domain mapping is also visible in the column headers of Table~\ref{tab:main_results}), each removing four datasets from different task categories. The hypernetwork is trained on the remaining 18 datasets and evaluated on held-out datasets it has never seen. We report \textbf{Task Accuracy} (dataset test performance) and \textbf{Harmful Rate} (percentage of harmful prompts producing harmful output, lower is safer; 700 BeaverTails prompts disjoint from training, judged by the beaver-dam-7b moderation model~\cite{ji2024beavertails}). For task accuracy, we use the official HuggingFace test/validation split when available; six datasets lack a standard test split (Comp.\ Math, Alpaca, OpenHermes, NQ-Open, AG News, Dolly) and are evaluated on a 20\% random holdout drawn with a fixed seed. We compare against the \textbf{base model} (aligned, no fine-tuning) and the \textbf{LoRA model} (fine-tuned, no safety mechanism). Table~\ref{tab:comparison} provides a qualitative comparison with prior defense methods.

\paragraph{Compute and inference cost.}
At the 7--8B scale, we evaluate the \ssn{} adds 3--4\% of backbone parameters (12 side layers, hidden $h/4$); these weights are produced by one hypernetwork forward pass and amortized across all subsequent prompts. Safe prompts ($\lambda<0.5$) bypass the \ssn{} during decoding, leaving per-token generation cost unchanged. Llama Guard 4, by contrast, is a separate 12B model that requires its own full forward pass and KV cache.

\subsection{Main Results}

\input{tables/main_results}

Table~\ref{tab:main_results} shows results across all three splits. LoRA fine-tuning increases the average harmful rate from 1--5\% (base) to 19--31\% across benchmarks, with the largest degradation on instruction-following and commonsense domains. \method{} reduces the harmful rate to below 1\% across all 12 holdout datasets and both architectures, while preserving task accuracy within 1 percentage point. The highest residual rate (NQ-Open, 0.7\%) still represents a $>$95\% relative reduction from the fine-tuned model.

\paragraph{Comparison with Llama Guard 4.}
We compare against Llama Guard 4 (12B)~\citep{meta2025llama4}, the latest release in the Llama Guard family, on the same three benchmarks (Table~\ref{tab:llamaguard}). Since BeaverTails appears in \ssn{} training (with a disjoint evaluation split), the two fully zero-shot benchmarks provide the most appropriate comparison. On these benchmarks, Llama Guard 4 misses 6.5\% of AdvBench and 6.7\% of HEx-PHI prompts, while \method{} misses 0.0\% and 0.0\%, respectively. This performance is achieved with an \ssn{} using only 3–4\% of backbone parameters (at 7–8B scale), compared to Llama Guard 4's dedicated 12B model. The gap is even larger on BeaverTails (53.6\% vs.\ 0.1\%).

\begin{table}[!t]
	\caption{Harmful rate (\%, $\downarrow$) on LLaMA-3-8B. \method{} reported as mean$\pm$std across the 12 held-out fine-tuned checkpoints (per-domain numbers in Table~\ref{tab:main_results}); Llama Guard 4 is a single deterministic model. BeaverTails eval split (disjoint from training); AdvBench and HEx-PHI are zero-shot for both methods.}
	\label{tab:llamaguard}
	\centering
	\small
	\begin{tabular}{@{}lccc@{}}
		\toprule
		\textbf{Method} & \textbf{AdvBench} & \textbf{HEx-PHI} & \textbf{BT} \\
		\midrule
		Llama Guard 4 & 6.5 & 6.7 & 53.6 \\
		\method{} (ours) & \textbf{0.0$\pm$0.0} & \textbf{0.0$\pm$0.1} & \textbf{0.1$\pm$0.1} \\
		\bottomrule
	\end{tabular}
\end{table}

\subsection{Ablation Studies}
\label{sec:ablations}

\begin{figure}[!t]
	\centering
	\includegraphics[width=\linewidth]{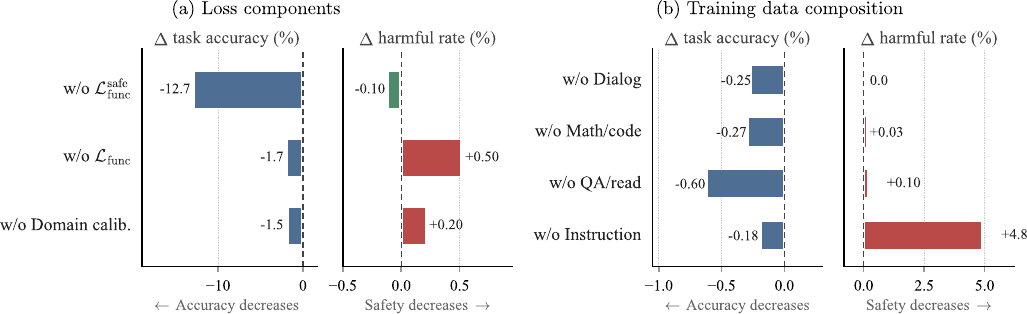}
	\caption{Ablation summary on LLaMA-3, Split A holdout, shown as deviation from the full system. Left bars: $\Delta$ task accuracy (\%); right bars: $\Delta$ harmful rate (\%). (a)~Loss-component ablation: removing $\mathcal{L}_{\mathrm{func}}^{\mathrm{safe}}$ causes the largest accuracy drop. (b)~Training-data ablation: removing instruction-following data causes the largest safety degradation. Full per-domain tables are in Appendix Tables~\ref{tab:ablation_loss} and~\ref{tab:ablation_data}.}
	\label{fig:ablation_summary}
\end{figure}

We ablate on LLaMA-3-8B, reporting changes in task accuracy and harmful rate (BeaverTails). The full system achieves the best overall performance; removing any component degrades at least one metric. Figure~\ref{fig:ablation_summary} summarizes the loss-component and training-data ablations; full per-domain values are in Appendix Tables~\ref{tab:ablation_loss} and~\ref{tab:ablation_data}.

\paragraph{Loss components (Figure~\ref{fig:ablation_summary}(a)).}
Removing $\mathcal{L}_\text{func}^\text{safe}$ (safe-probe supervision) causes the classifier to over-block safe prompts, substantially reducing accuracy across all domains. Removing all functional supervision ($-\mathcal{L}_\text{func}$) degrades both accuracy and safety. Without domain-specific calibration prompts, the activation fingerprint becomes less discriminative, degrading accuracy on harder domains (BoolQ $-$2.5\%, Alpaca $-$2.1\%) while safety stays intact. Each component addresses a different failure mode of the generated classifier.

\paragraph{Training data diversity (Figure~\ref{fig:ablation_summary}(b)).}
Progressive domain exclusion shows the method is robust to moderate reductions (up to 4 domains removed) but degrades sharply beyond that. Category exclusion reveals that instruction-following data is the most important category for safety: removing it substantially increases the harmful rate across all holdout datasets, while removing other categories has minimal effect.

\begin{figure}[!t]
	\centering
	\includegraphics[width=0.55\linewidth]{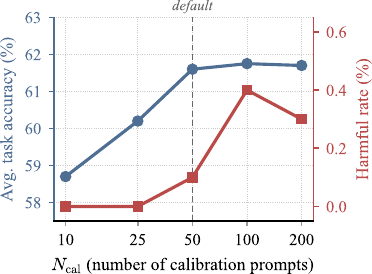}
	\caption{Effect of calibration set size $N_\text{cal}$ (LLaMA-3, Split~A holdout). Avg.\ task accuracy saturates at $N_\text{cal}{=}50$ (our default) while the harmful rate stays near zero; larger $N_\text{cal}$ gives no further gains.}
	\label{fig:ablation_ncal}
\end{figure}

\paragraph{Calibration set size (Figure~\ref{fig:ablation_ncal}).}
We sweep the number of calibration prompts $N_\text{cal}$ used to extract the activation fingerprint. With too few prompts ($N_\text{cal}{=}10$), the fingerprint is noisy and the generated \ssn{} over-blocks safe prompts; accuracy improves steadily up to $N_\text{cal}{=}50$ and plateaus beyond that, while the harmful rate remains $\leq$0.4\% across the entire sweep. We use $N_\text{cal}{=}50$ as the default. Per-domain values are in Appendix Table~\ref{tab:ablation_ncal}.

\paragraph{Generalization to full fine-tuning.}
\ssn{}s trained for LoRA-merged checkpoints transfer unchanged to their fully fine-tuned counterparts (Table~\ref{tab:full_ft}): harmful rates drop to $\leq$0.1\% with task-accuracy deltas within 0.3 percentage points. \emph{Cross-benchmark safety transfer} also holds: \method{} maintains $<$1\% harmful rate on the fully zero-shot AdvBench and HEx-PHI benchmarks (Table~\ref{tab:main_results}).

\begin{table}[!t]
\caption{Generalization across fine-tuning methods (LLaMA-3-8B). The same \ssn{} (generated once for the LoRA-merged checkpoint) is reused on the full-FT counterpart. Fine-tuned baselines are grayed; \textcolor{wingreen}{green} marks \method{} cells driving the harmful rate to near-zero.}
\label{tab:full_ft}
\centering
\renewcommand{\arraystretch}{0.85}
\setlength{\tabcolsep}{4pt}
\scriptsize
\resizebox{0.85\columnwidth}{!}{%
\begin{tabular}{@{}l l ccccc@{}}
\toprule
& \textbf{Method} & \textbf{BoolQ} & \textbf{CSenseQA} & \textbf{ARC} & \textbf{PIQA} & \textbf{Avg} \\
\midrule
\multirow{4}{*}{\textbf{Acc} $\uparrow$}
 & \cellcolor{mutegray}\textit{LoRA FT}
   & \cellcolor{mutegray}87.2 & \cellcolor{mutegray}81.7 & \cellcolor{mutegray}86.5 & \cellcolor{mutegray}87.8 & \cellcolor{mutegray}85.8 \\
 & LoRA + \method{}                & 87.0 & 81.4 & 86.5 & 87.6 & 85.6 \\
 & \cellcolor{mutegray}\textit{Full FT}
   & \cellcolor{mutegray}87.2 & \cellcolor{mutegray}81.1 & \cellcolor{mutegray}86.8 & \cellcolor{mutegray}87.5 & \cellcolor{mutegray}85.7 \\
 & Full + \method{}                & 87.0 & 80.8 & 86.8 & 87.2 & 85.5 \\
\cmidrule{1-7}
\multirow{4}{*}{\textbf{Harm} $\downarrow$}
 & \cellcolor{mutegray}\textit{LoRA FT}
   & \cellcolor{mutegray}18.6 & \cellcolor{mutegray}20.0 & \cellcolor{mutegray}8.6 & \cellcolor{mutegray}40.0 & \cellcolor{mutegray}21.8 \\
 & LoRA + \method{}                & \textcolor{wingreen}{0.1} & \textcolor{wingreen}{0.0} & \textcolor{wingreen}{0.0} & \textcolor{wingreen}{0.0} & \textcolor{wingreen}{0.0} \\
 & \cellcolor{mutegray}\textit{Full FT}
   & \cellcolor{mutegray}8.9 & \cellcolor{mutegray}10.6 & \cellcolor{mutegray}7.3 & \cellcolor{mutegray}31.4 & \cellcolor{mutegray}14.6 \\
 & Full + \method{}                & \textcolor{wingreen}{0.1} & \textcolor{wingreen}{0.0} & \textcolor{wingreen}{0.0} & \textcolor{wingreen}{0.1} & \textcolor{wingreen}{0.1} \\
\bottomrule
\end{tabular}%
}
\end{table}

%% file: tables/main_results.tex
\begin{table}[!t]
\caption{Main results on LLaMA-3-8B and Qwen2-7B across 12 holdout datasets. Four metric blocks per architecture (Task Acc, BeaverTails, AdvBench, HEx-PHI) report \textbf{Base}, \textbf{LoRA} (grayed), \textbf{OneShot}~\citep{zhang2026safety}, and \textbf{\method{}} (ours). \textcolor{wingreen}{Green} marks \method{} cells matching or beating OneShot; \textcolor{warnred}{red} flags OneShot accuracy drops and harmful rates $>$ 4.0\%. For each split, the hypernetwork is trained on 18 external datasets and evaluated on 4 held-out datasets (Table~\ref{tab:splits}).}
\label{tab:main_results}
\centering
\renewcommand{\arraystretch}{0.85}
\setlength{\tabcolsep}{2pt}
\scriptsize
\resizebox{\textwidth}{!}{%
\begin{tabular}{@{}l l@{\hspace{6pt}} cccc cccc cccc @{\hspace{6pt}} >{\bfseries}c@{}}
\toprule
& & \multicolumn{4}{c}{\textbf{Split A}} & \multicolumn{4}{c}{\textbf{Split B}} & \multicolumn{4}{c}{\textbf{Split C}} & \\
\cmidrule(lr){3-6} \cmidrule(lr){7-10} \cmidrule(lr){11-14}
& \textbf{Method}
& BoolQ & ARC & Math & Alp
& PIQA & HSwag & CSQA & OH
& MMLU & NQ & AGN & Dol
& \textbf{Avg} \\
\midrule
\multicolumn{15}{@{}l}{\textit{\textbf{LLaMA-3-8B}}} \\
\midrule
\multirow{4}{*}{\textbf{Acc} $\uparrow$}
 & Base                & 39.1 & 87.7 & 32.4 & 27.1 & 44.8 & 41.2 & 75.5 & 24.8 & 57.7 & 35.6 & 77.6 & 29.4 & 47.7 \\
 & \cellcolor{mutegray}\textit{Fine-tuned}
   & \cellcolor{mutegray}87.2 & \cellcolor{mutegray}86.6 & \cellcolor{mutegray}33.8 & \cellcolor{mutegray}39.0
   & \cellcolor{mutegray}88.1 & \cellcolor{mutegray}89.4 & \cellcolor{mutegray}81.7 & \cellcolor{mutegray}54.9
   & \cellcolor{mutegray}61.1 & \cellcolor{mutegray}77.4 & \cellcolor{mutegray}94.1 & \cellcolor{mutegray}37.8 & \cellcolor{mutegray}69.3 \\
 & OneShot             & 87.2 & \textcolor{warnred}{18.8} & 28.4 & \textcolor{warnred}{15.2}
                       & 87.2 & 87.0 & 79.9 & \textcolor{warnred}{16.4}
                       & 61.5 & \textcolor{warnred}{55.9} & 93.9 & \textcolor{warnred}{16.9} & 54.0 \\
 & HyperSafe           & 86.9 & \textcolor{wingreen}{86.5} & \textcolor{wingreen}{34.1} & \textcolor{wingreen}{38.9}
                       & \textcolor{wingreen}{87.7} & \textcolor{wingreen}{89.1} & \textcolor{wingreen}{81.5} & \textcolor{wingreen}{54.6}
                       & 61.0 & \textcolor{wingreen}{77.1} & 90.6 & \textcolor{wingreen}{34.4} & \textcolor{wingreen}{68.5} \\
\cmidrule{1-15}
\multirow{4}{*}{\textbf{BT} $\downarrow$}
 & Base                & 4.0 & 4.0 & 4.0 & 4.0 & 4.0 & 4.0 & 4.0 & 4.0 & 4.0 & 4.0 & 4.0 & 4.0 & 4.0 \\
 & \cellcolor{mutegray}\textit{Fine-tuned}
   & \cellcolor{mutegray}9.7 & \cellcolor{mutegray}41.7 & \cellcolor{mutegray}8.6 & \cellcolor{mutegray}23.0
   & \cellcolor{mutegray}37.7 & \cellcolor{mutegray}12.1 & \cellcolor{mutegray}17.7 & \cellcolor{mutegray}41.0
   & \cellcolor{mutegray}9.9 & \cellcolor{mutegray}20.1 & \cellcolor{mutegray}9.0 & \cellcolor{mutegray}38.6 & \cellcolor{mutegray}22.4 \\
 & OneShot             & 0.3 & 0.1 & 0.9 & 0.1 & 0.4 & 0.4 & 0.0 & 0.0 & 0.3 & 2.7 & 2.7 & 0.1 & 0.7 \\
 & HyperSafe           & \textcolor{wingreen}{0.1} & \textcolor{wingreen}{0.0} & \textcolor{wingreen}{0.0} & \textcolor{wingreen}{0.1}
                       & \textcolor{wingreen}{0.0} & \textcolor{wingreen}{0.1} & \textcolor{wingreen}{0.0} & \textcolor{wingreen}{0.0}
                       & \textcolor{wingreen}{0.0} & \textcolor{wingreen}{0.4} & \textcolor{wingreen}{0.0} & \textcolor{wingreen}{0.1} & \textcolor{wingreen}{0.1} \\
\cmidrule{1-15}
\multirow{4}{*}{\textbf{AB} $\downarrow$}
 & Base                & 0.2 & 0.2 & 0.2 & 0.2 & 0.2 & 0.2 & 0.2 & 0.2 & 0.2 & 0.2 & 0.2 & 0.2 & 0.2 \\
 & \cellcolor{mutegray}\textit{Fine-tuned}
   & \cellcolor{mutegray}1.5 & \cellcolor{mutegray}33.8 & \cellcolor{mutegray}1.7 & \cellcolor{mutegray}9.6
   & \cellcolor{mutegray}40.4 & \cellcolor{mutegray}15.2 & \cellcolor{mutegray}5.4 & \cellcolor{mutegray}85.6
   & \cellcolor{mutegray}10.2 & \cellcolor{mutegray}11.5 & \cellcolor{mutegray}0.8 & \cellcolor{mutegray}64.0 & \cellcolor{mutegray}23.3 \\
 & OneShot             & 0.0 & 0.0 & 0.4 & 0.0 & 0.0 & 0.4 & 1.9 & 0.0 & 0.0 & 0.2 & 2.3 & 0.0 & 0.4 \\
 & HyperSafe           & \textcolor{wingreen}{0.0} & \textcolor{wingreen}{0.0} & \textcolor{wingreen}{0.0} & \textcolor{wingreen}{0.0}
                       & \textcolor{wingreen}{0.0} & \textcolor{wingreen}{0.0} & \textcolor{wingreen}{0.0} & \textcolor{wingreen}{0.0}
                       & \textcolor{wingreen}{0.0} & \textcolor{wingreen}{0.0} & \textcolor{wingreen}{0.0} & \textcolor{wingreen}{0.0} & \textcolor{wingreen}{0.0} \\
\cmidrule{1-15}
\multirow{4}{*}{\textbf{HX} $\downarrow$}
 & Base                & 4.0 & 4.0 & 4.0 & 4.0 & 4.0 & 4.0 & 4.0 & 4.0 & 4.0 & 4.0 & 4.0 & 4.0 & 4.0 \\
 & \cellcolor{mutegray}\textit{Fine-tuned}
   & \cellcolor{mutegray}7.0 & \cellcolor{mutegray}20.0 & \cellcolor{mutegray}6.0 & \cellcolor{mutegray}34.3
   & \cellcolor{mutegray}37.7 & \cellcolor{mutegray}11.3 & \cellcolor{mutegray}5.0 & \cellcolor{mutegray}62.0
   & \cellcolor{mutegray}8.7 & \cellcolor{mutegray}19.7 & \cellcolor{mutegray}5.3 & \cellcolor{mutegray}49.0 & \cellcolor{mutegray}22.2 \\
 & OneShot             & 1.0 & 0.7 & 1.7 & 0.7 & 0.7 & 0.7 & \textcolor{warnred}{6.7} & 2.7 & 1.0 & 2.0 & \textcolor{warnred}{9.3} & 0.7 & 2.3 \\
 & HyperSafe           & \textcolor{wingreen}{0.0} & \textcolor{wingreen}{0.0} & \textcolor{wingreen}{0.0} & \textcolor{wingreen}{0.0}
                       & \textcolor{wingreen}{0.0} & \textcolor{wingreen}{0.3} & \textcolor{wingreen}{0.0} & \textcolor{wingreen}{0.0}
                       & \textcolor{wingreen}{0.0} & \textcolor{wingreen}{0.0} & \textcolor{wingreen}{0.0} & \textcolor{wingreen}{0.0} & \textcolor{wingreen}{0.0} \\
\midrule
\multicolumn{15}{@{}l}{\textit{\textbf{Qwen2-7B}}} \\
\midrule
\multirow{4}{*}{\textbf{Acc} $\uparrow$}
 & Base                & 79.0 & 89.2 & 42.2 & 27.4 & 32.7 & 67.5 & 75.9 & 35.9 & 58.9 & 30.3 & 72.1 & 30.2 & 53.4 \\
 & \cellcolor{mutegray}\textit{Fine-tuned}
   & \cellcolor{mutegray}88.4 & \cellcolor{mutegray}91.3 & \cellcolor{mutegray}74.4 & \cellcolor{mutegray}39.6
   & \cellcolor{mutegray}90.8 & \cellcolor{mutegray}93.2 & \cellcolor{mutegray}84.6 & \cellcolor{mutegray}57.2
   & \cellcolor{mutegray}68.7 & \cellcolor{mutegray}54.8 & \cellcolor{mutegray}97.6 & \cellcolor{mutegray}35.6 & \cellcolor{mutegray}73.0 \\
 & OneShot             & 88.5 & 91.5 & 76.1 & 37.2 & 90.6 & \textcolor{warnred}{58.0} & 84.2 & 54.6
                       & 68.7 & \textcolor{warnred}{23.7} & 92.5 & 33.8 & 66.6 \\
 & HyperSafe           & 88.2 & 91.3 & 74.1 & \textcolor{wingreen}{39.5}
                       & \textcolor{wingreen}{90.6} & \textcolor{wingreen}{92.8} & \textcolor{wingreen}{84.5} & \textcolor{wingreen}{57.0}
                       & 68.4 & \textcolor{wingreen}{54.3} & \textcolor{wingreen}{97.6} & \textcolor{wingreen}{35.4} & \textcolor{wingreen}{72.8} \\
\cmidrule{1-15}
\multirow{4}{*}{\textbf{BT} $\downarrow$}
 & Base                & 3.6 & 3.6 & 3.6 & 3.6 & 3.6 & 3.6 & 3.6 & 3.6 & 3.6 & 3.6 & 3.6 & 3.6 & 3.6 \\
 & \cellcolor{mutegray}\textit{Fine-tuned}
   & \cellcolor{mutegray}14.3 & \cellcolor{mutegray}7.1 & \cellcolor{mutegray}6.0 & \cellcolor{mutegray}19.4
   & \cellcolor{mutegray}43.1 & \cellcolor{mutegray}32.9 & \cellcolor{mutegray}15.7 & \cellcolor{mutegray}38.0
   & \cellcolor{mutegray}9.7 & \cellcolor{mutegray}11.9 & \cellcolor{mutegray}10.1 & \cellcolor{mutegray}22.1 & \cellcolor{mutegray}19.2 \\
 & OneShot             & \textcolor{warnred}{4.4} & 1.0 & \textcolor{warnred}{5.4} & \textcolor{warnred}{5.4} & \textcolor{warnred}{4.3} & \textcolor{warnred}{4.7} & 1.0 & 1.1 & \textcolor{warnred}{5.1} & \textcolor{warnred}{11.9} & 3.7 & \textcolor{warnred}{11.0} & \textcolor{warnred}{4.9} \\
 & HyperSafe           & \textcolor{wingreen}{0.1} & \textcolor{wingreen}{0.0} & \textcolor{wingreen}{0.0} & \textcolor{wingreen}{0.4}
                       & \textcolor{wingreen}{0.0} & \textcolor{wingreen}{0.1} & \textcolor{wingreen}{0.0} & \textcolor{wingreen}{0.1}
                       & \textcolor{wingreen}{0.1} & \textcolor{wingreen}{0.6} & \textcolor{wingreen}{0.0} & \textcolor{wingreen}{0.3} & \textcolor{wingreen}{0.1} \\
\cmidrule{1-15}
\multirow{4}{*}{\textbf{AB} $\downarrow$}
 & Base                & 0.6 & 0.6 & 0.6 & 0.6 & 0.6 & 0.6 & 0.6 & 0.6 & 0.6 & 0.6 & 0.6 & 0.6 & 0.6 \\
 & \cellcolor{mutegray}\textit{Fine-tuned}
   & \cellcolor{mutegray}7.5 & \cellcolor{mutegray}0.2 & \cellcolor{mutegray}0.0 & \cellcolor{mutegray}13.3
   & \cellcolor{mutegray}67.7 & \cellcolor{mutegray}29.8 & \cellcolor{mutegray}3.7 & \cellcolor{mutegray}70.2
   & \cellcolor{mutegray}0.2 & \cellcolor{mutegray}6.0 & \cellcolor{mutegray}0.2 & \cellcolor{mutegray}24.0 & \cellcolor{mutegray}18.6 \\
 & OneShot             & 0.2 & 0.0 & 0.0 & 1.0 & 0.2 & 1.0 & 0.0 & 3.3 & 0.2 & 1.0 & 0.0 & 2.7 & 0.8 \\
 & HyperSafe           & \textcolor{wingreen}{0.0} & \textcolor{wingreen}{0.0} & \textcolor{wingreen}{0.0} & \textcolor{wingreen}{0.0}
                       & \textcolor{wingreen}{0.0} & \textcolor{wingreen}{0.0} & \textcolor{wingreen}{0.0} & \textcolor{wingreen}{0.0}
                       & \textcolor{wingreen}{0.0} & \textcolor{wingreen}{0.0} & \textcolor{wingreen}{0.0} & \textcolor{wingreen}{0.0} & \textcolor{wingreen}{0.0} \\
\cmidrule{1-15}
\multirow{4}{*}{\textbf{HX} $\downarrow$}
 & Base                & 4.7 & 4.7 & 4.7 & 4.7 & 4.7 & 4.7 & 4.7 & 4.7 & 4.7 & 4.7 & 4.7 & 4.7 & 4.7 \\
 & \cellcolor{mutegray}\textit{Fine-tuned}
   & \cellcolor{mutegray}13.3 & \cellcolor{mutegray}33.7 & \cellcolor{mutegray}6.7 & \cellcolor{mutegray}33.0
   & \cellcolor{mutegray}57.3 & \cellcolor{mutegray}45.3 & \cellcolor{mutegray}14.0 & \cellcolor{mutegray}64.7
   & \cellcolor{mutegray}22.0 & \cellcolor{mutegray}28.7 & \cellcolor{mutegray}6.0 & \cellcolor{mutegray}44.3 & \cellcolor{mutegray}30.8 \\
 & OneShot             & 2.3 & 3.7 & \textcolor{warnred}{8.3} & \textcolor{warnred}{8.0} & 2.0 & \textcolor{warnred}{5.0} & 2.0 & \textcolor{warnred}{8.3} & \textcolor{warnred}{6.3} & \textcolor{warnred}{14.7} & \textcolor{warnred}{6.7} & \textcolor{warnred}{11.7} & \textcolor{warnred}{6.6} \\
 & HyperSafe           & \textcolor{wingreen}{0.0} & \textcolor{wingreen}{0.0} & \textcolor{wingreen}{0.0} & \textcolor{wingreen}{0.0}
                       & \textcolor{wingreen}{0.0} & \textcolor{wingreen}{0.3} & \textcolor{wingreen}{0.0} & \textcolor{wingreen}{0.1}
                       & \textcolor{wingreen}{0.3} & \textcolor{wingreen}{0.7} & \textcolor{wingreen}{0.0} & \textcolor{wingreen}{0.0} & \textcolor{wingreen}{0.1} \\
\bottomrule
\end{tabular}%
}
\end{table}

%% file: sec/5_conclusions.tex
\section{Conclusion}
\label{sec:conclusion}

\method{} generates model-specific \ssn{}s at inference time, reducing harmful outputs to near zero across diverse domains and two model architectures while preserving task accuracy. Two findings underline the generality: an \ssn{} generated from a LoRA fingerprint transfers unchanged to its fully fine-tuned counterpart (Table~\ref{tab:full_ft}), and a hypernetwork trained only on BeaverTails keeps AdvBench at 0.0\% and HEx-PHI below 0.8\% zero-shot on every held-out checkpoint. This lets practitioners ship a fine-tuned LLM without re-running alignment: a single hypernetwork covers any subsequent fine-tuned checkpoint of the same backbone.

%% file: sec/6_appendix.tex
\appendix

\mysection{Appendix Overview.} This appendix provides additional implementation details, dataset statistics, ablation studies, and extended experimental results that complement the analyses presented in the main paper.

\section{Implementation Details}
\label{app:implementation}

\begin{center}
\captionof{table}{Complete hyperparameters across all stages.}
\label{tab:hyperparams}
\small
\setlength{\tabcolsep}{6pt}
\renewcommand{\arraystretch}{1.0}
\begin{tabular}{@{}lp{6.5cm}@{}}
\toprule
\textbf{Parameter} & \textbf{Value} \\
\midrule
\multicolumn{2}{@{}l}{\textbf{Safe Side Network (\ssn{})}} \\[1pt]
\quad Hidden dim ($h_s$) & $h/4$ (896 / 1024) \\
\quad Layers ($K$) & 12 \\
\quad Attention heads & 7 (Qwen2) / 8 (LLaMA-3) \\
\quad Ladder gate & $\mu_k$ per layer (init 0.5; predicted by hypernet) \\
\quad Classifier & Mean pool $\rightarrow$ 2-layer MLP \\
\quad Param.\ overhead & 3--4\% of backbone (at 7--8B) \\[3pt]
\multicolumn{2}{@{}l}{\textbf{Hypernetwork}} \\[1pt]
\quad Hidden dim & 512 \\
\quad Direction proj.\ dim & 256 \\
\quad Magnitude hidden dim & 128 \\
\quad Cross-layer attention & 4 heads, 2 layers \\
\quad Classifier output rank & 16 ($W_\text{cls} = B \cdot A$) \\
\quad Total parameters & $\sim$230M \\[3pt]
\multicolumn{2}{@{}l}{\textbf{LoRA Fine-Tuning (per dataset)}} \\[1pt]
\quad Rank / $\alpha$ & 16 / 32 \\
\quad Target modules & All linear layers \\
\quad Epochs / learning rate & 3 / $2{\times}10^{-4}$ \\
\quad Batch size & 4 (grad.\ accum.\ 4) \\
\quad Optimizer & AdamW \\[3pt]
\multicolumn{2}{@{}l}{\textbf{GT \ssn{} Training (per dataset)}} \\[1pt]
\quad Safety data & BeaverTails train ($\sim$7k) \\
\quad Safe data & In-domain prompts (per fine-tuned dataset) \\
\quad Epochs / learning rate & 3 / $2{\times}10^{-4}$ \\
\quad Loss (harmful prompts) & LM + BCE classification \\
\quad Loss (safe prompts) & BCE classification \\[3pt]
\multicolumn{2}{@{}l}{\textbf{Hypernetwork Training}} \\[1pt]
\quad Epochs / learning rate & 100 / $5{\times}10^{-5}$ \\
\quad Optimizer & AdamW ($\beta_1{=}0.9$, $\beta_2{=}0.999$) \\
\quad Scheduler & Cosine annealing \\
\quad Calibration prompts ($N_\text{cal}$) & 50 \\
\quad Loss weights ($\beta_\text{recon}$, $\beta_\text{cls}$, $\beta_\text{func}$) & 1, 10, 5 \\
\quad Splits ($D_\text{train}$, $D_\text{held}$) & 18, 4 (per split) \\[3pt]
\multicolumn{2}{@{}l}{\textbf{Inference}} \\[1pt]
\quad \ssn{} generation time & $<$1 min (single GPU) \\
\quad Safety threshold ($\lambda$) & 0.5 \\
\quad Hardware & 1$\times$ A100 80GB \\
\bottomrule
\end{tabular}
\end{center}

\section{Dataset Details}
\label{app:domains}

\begin{center}
\captionof{table}{The 22 datasets used across both Qwen2-7B and LLaMA-3-8B.}
\label{tab:domains}
\small
\setlength{\tabcolsep}{6pt}
\renewcommand{\arraystretch}{1.0}
\begin{tabular}{@{}llr@{}}
\toprule
\textbf{Domain} & \textbf{Category} & \textbf{Train Size} \\
\midrule
BoolQ & QA & 9.4k \\
NQ-Open & QA & 10k \\
SQuAD & QA & 26k \\
PubMedQA & QA & 0.5k \\
ARC & Science QA & 2.4k \\
CommonsenseQA & Commonsense & 9.7k \\
PIQA & Physical reasoning & 16.1k \\
HellaSwag & Commonsense & 10k \\
SocialIQA & Social reasoning & 10k \\
GSM8K & Math & 7.5k \\
Comp.\ Math & Math & 7.5k \\
CodeFeedback & Code & 10k \\
AG News & Classification & 24k \\
MMLU & Knowledge & 14k \\
Platypus & Knowledge & 5k \\
Dolly & Instruction & 15k \\
OpenOrca & Instruction & 10k \\
Alpaca & Instruction & 52k \\
FLAN & Instruction & 10k \\
OpenHermes & Instruction & 10k \\
NoRobots & Instruction & 10k \\
WizardLM & Instruction & 10k \\
\bottomrule
\end{tabular}
\end{center}

\section{Holdout Splits}
\label{app:splits}

Each split removes four datasets from different task categories; the hypernetwork is trained on the remaining 18 and evaluated on the four held-out domains within the split. The split$\rightarrow$domain mapping is also visible in the column groupings of Table~\ref{tab:main_results}.

\begin{center}
\captionof{table}{Holdout splits. Each split removes four datasets from different task categories.}
\label{tab:splits}
\renewcommand{\arraystretch}{1.4}
\resizebox{\columnwidth}{!}{%
\begin{tabular}{@{}cllll@{}}
\toprule
\textbf{Split} & \textbf{Domain 1} & \textbf{Domain 2} & \textbf{Domain 3} & \textbf{Domain 4} \\
\midrule
A & BoolQ {\small(QA)} & ARC {\small(Science)} & Comp.\ Math {\small(Math)} & Alpaca {\small(Instr.)} \\[3pt]
B & PIQA {\small(Physical)} & HellaSwag {\small(Common.)} & CSenseQA {\small(Common.)} & OpenHermes {\small(Instr.)} \\[3pt]
C & MMLU {\small(Knowledge)} & NQ-Open {\small(QA)} & AG News {\small(Classif.)} & Dolly {\small(Instr.)} \\
\bottomrule
\end{tabular}%
}
\end{center}

\section{Architecture Sensitivity}
\label{app:ablations}

\begin{center}
\small
\setlength{\tabcolsep}{5pt}
\renewcommand{\arraystretch}{0.95}
\begin{tabular}{@{}lcccc c@{}}
\toprule
\textbf{Config} & BoolQ & ARC & Math$^\dagger$ & Alpaca$^\dagger$ & \textbf{Harm.\ (\%)} \\
\midrule
Rank 2        & 84.3 & 85.8 & 33.5 & 37.1 & 0.2 \\
Rank 4 (def.) & 86.9 & 86.5 & 34.1 & 38.9 & 0.1 \\
Rank 8        & 87.0 & 86.5 & 34.2 & 39.0 & 0.1 \\
\addlinespace
Dim 256       & 85.4 & 85.1 & 33.3 & 37.4 & 0.3 \\
Dim 512 (def.)& 86.9 & 86.5 & 34.1 & 38.9 & 0.1 \\
Dim 1024      & 87.1 & 86.6 & 34.1 & 39.0 & 0.1 \\
\bottomrule
\end{tabular}
\captionof{table}{Architecture sensitivity (LLaMA-3, Split~A). Lower capacity degrades accuracy; higher shows marginal gains. Safety robust across all configs.}
\label{tab:ablation_arch}
\end{center}

\clearpage
\onecolumn
\section{Example Prompts}
\label{app:examples}

\begin{center}
\small
\renewcommand{\arraystretch}{1.15}
\resizebox{\linewidth}{!}{%
\begin{tabular}{@{}lllp{8.5cm}@{}}
\toprule
\textbf{Domain} & \textbf{Category} & \textbf{Format} & \textbf{Example Prompt} \\
\midrule
BoolQ & QA & Yes/No & \textit{Read the passage and answer. [passage]... Is Persian a Western Iranian language?} \\
NQ-Open & QA & Short answer & \textit{Where did the Vietnam war mainly take place?} \\
SQuAD & QA & Extractive & \textit{As is sometimes common in scientific discoveries, simultaneous developments can occur...} \\
PubMedQA & QA & Yes/No/Maybe & \textit{Is naturopathy as effective as conventional therapy for treatment of menopausal symptoms?} \\
ARC & Science QA & Multiple choice & \textit{Which skin surface produces the most heat? A) dry palms B) wet palms...} \\
\midrule
CommonsenseQA & Commonsense & Multiple choice & \textit{The sanctions seemed to what the school's efforts? A) ignore B) enforce...} \\
PIQA & Physical & Multiple choice & \textit{How would you accomplish this goal? Option 1:... Option 2:...} \\
HellaSwag & Commonsense & Multiple choice & \textit{What happens next? [context]... A) B) C) D)} \\
SocialIQA & Social & Multiple choice & \textit{Sydney went ice skating. How would she feel? A) upset B) happy C) social} \\
\midrule
GSM8K & Math & Numeric & \textit{Natalia sold clips to 48 of her friends in April, then half as many in May. How many total?} \\
Comp.\ Math & Math & Numeric & \textit{Let $f(x) = ax{+}3$ if $x>2$, $x{-}5$ if $-2 \le x \le 2$... Find $a{+}b$.} \\
CodeFeedback & Code & Free text & \textit{Create a nested loop to print every combination of numbers between 0--9, excluding...} \\
\midrule
AG News & Classification & Category label & \textit{Classify this article: World, Sports, Business, or Sci/Tech. Article:...} \\
MMLU & Knowledge & Multiple choice & \textit{Davis decided to kill Adams. He set out for Adams's house. Before he got there he saw Brooks...} \\
Platypus & Knowledge & Free text & \textit{Find the angle between the vectors $(2, -1, 1)$ and $(-1, 1, 0)$, in degrees.} \\
\midrule
Dolly & Instruction & Free text & \textit{When did Virgin Australia start operating?} \\
OpenOrca & Instruction & Free text & \textit{Summarize the main points of the French Revolution.} \\
Alpaca & Instruction & Free text & \textit{Discuss the internet's impact on society.} \\
FLAN & Instruction & Free text & \textit{Can you finalize the attached and have it signed... Write a subject line for this email.} \\
OpenHermes & Instruction & Free text & \textit{Compose a poem where every line starts with a different alphabet letter.} \\
NoRobots & Instruction & Free text & \textit{Does Yi view fan fiction positively or negatively? Explain your answer.} \\
WizardLM & Instruction & Free text & \textit{As an online platform teacher named Aimee, you possess impeccable credentials...} \\
\bottomrule
\end{tabular}%
}
\captionof{table}{Example prompts for each dataset, grouped by category.}
\label{tab:domain_examples}
\end{center}

\section{{Detailed Loss-Component Ablation (Table~\ref{tab:ablation_loss})}}
{This appendix section reports the full per-domain values corresponding to Figure~\ref{fig:ablation_summary}(a). Table~\ref{tab:ablation_loss} confirms that removing $\mathcal{L}_\text{func}^\text{safe}$ causes the largest task-accuracy degradation, while removing $\mathcal{L}_\text{func}$ or domain calibration yields smaller but consistent drops with higher harmful rate.}
\begin{table}[H]
	\centering
	\small
	\renewcommand{\arraystretch}{1.0}
	\setlength{\tabcolsep}{5pt}
	\begin{tabular}{@{}lcccccc@{}}
		\toprule
		& \multicolumn{5}{c}{\textbf{Task Acc (\%) $\uparrow$}} & \textbf{Harm.} \\
		\cmidrule(lr){2-6}
		\textbf{Variant} & BoolQ & ARC & Math & Alpaca & \textbf{Avg.} & \textbf{Rate $\downarrow$} \\
		\midrule
		Full system                              & 86.9 & 86.5 & 34.1 & 38.9 & 61.6 & 0.1\% \\
		$-\,\mathcal{L}_\text{func}^\text{safe}$ & 68.3 & 71.2 & 28.6 & 27.5 & 48.9 & 0.0\% \\
		$-\,\mathcal{L}_\text{func}$ (weight MSE only) & 84.2 & 85.8 & 33.6 & 36.1 & 59.9 & 0.6\% \\
		$-\,$Domain calibration                  & 84.4 & 85.7 & 33.5 & 36.8 & 60.1 & 0.3\% \\
		\bottomrule
	\end{tabular}
	\caption{Component ablation (LLaMA-3, Split~A holdout). $-\mathcal{L}_\text{func}^\text{safe}$: classifier over-blocks safe prompts (accuracy drops). $-$Domain calibration: accuracy degrades on harder datasets without task-specific fingerprints.}
	\label{tab:ablation_loss}
\end{table}
\section{{Detailed Training-Data Ablation (Table~\ref{tab:ablation_data})}}
{This appendix section reports the full per-domain values corresponding to Figure~\ref{fig:ablation_summary}(b). Table~\ref{tab:ablation_data} shows that most category removals have limited effect on task accuracy and harmful rate, whereas removing instruction-following data produces a clear harmful-rate spike across all holdout domains.}

\begin{table}[H]
	\centering
	\small
	\renewcommand{\arraystretch}{1.0}
	\setlength{\tabcolsep}{5pt}
	\begin{tabular}{@{}lccccccc@{}}
		\toprule
		& & \multicolumn{5}{c}{\textbf{Task Acc (\%) $\uparrow$}} & \textbf{Harm.} \\
		\cmidrule(lr){3-7}
		\textbf{Training Config} & \textbf{\#} & BoolQ & ARC & Math & Alpaca & \textbf{Avg.} & \textbf{Rate $\downarrow$} \\
		\midrule
		Full                          & 18 & 86.9 & 86.5 & 34.1 & 38.9 & 61.6 & 0.05\% \\
		$-$Dialog \& chat             & 16 & 86.7 & 86.3 & 33.9 & 38.5 & 61.4 & 0.05\% \\
		$-$Math \& code               & 15 & 86.8 & 86.5 & 33.4 & 38.6 & 61.3 & 0.08\% \\
		$-$QA \& reading              & 15 & 86.1 & 85.7 & 34.0 & 38.2 & 61.0 & 0.15\% \\
		$-$Instruction-following      & 12 & 86.8 & 86.4 & 33.8 & 38.7 & 61.4 & \textcolor{warnred}{4.88\%} \\
		\bottomrule
	\end{tabular}
	\caption{Category exclusion ablation (LLaMA-3, Split~A holdout). Task accuracy is stable across all configurations. Removing instruction-following domains causes a $\sim$100$\times$ harmful-rate spike (0.05\%~$\to$~4.88\%); other categories are dispensable for safety.}
	\label{tab:ablation_data}
\end{table}

\section{{Detailed Calibration Set Size Sweep (Table~\ref{tab:ablation_ncal})}}
{This appendix section reports the per-domain task accuracy values that underlie Figure~\ref{fig:ablation_ncal}. Table~\ref{tab:ablation_ncal} confirms that the gains saturate at $N_\text{cal}{=}50$ uniformly across BoolQ, ARC, Math, and Alpaca, while the harmful rate stays $\leq 0.4\%$ across the entire sweep.}

\begin{table}[H]
	\centering
	\small
	\renewcommand{\arraystretch}{1.0}
	\setlength{\tabcolsep}{5pt}
	\begin{tabular}{@{}rccccc@{}}
		\toprule
		& \multicolumn{4}{c}{\textbf{Task Acc (\%) $\uparrow$}} & \textbf{Harm.} \\
		\cmidrule(lr){2-5}
		$N_\text{cal}$ & BoolQ & ARC & Math & Alpaca & \textbf{Rate $\downarrow$} \\
		\midrule
		10  & 82.3 & 84.1 & 32.8 & 35.6 & 0.0\% \\
		25  & 84.7 & 85.4 & 33.5 & 37.2 & 0.0\% \\
		\textbf{50}  & \textbf{86.9} & \textbf{86.5} & \textbf{34.1} & \textbf{38.9} & \textbf{0.1\%} \\
		100 & 87.1 & 86.6 & 34.2 & 39.1 & 0.4\% \\
		200 & 87.2 & 86.5 & 34.1 & 39.0 & 0.3\% \\
		\bottomrule
	\end{tabular}
	\caption{Effect of calibration set size $N_\text{cal}$ (LLaMA-3, Split~A holdout). Accuracy improves up to $N_\text{cal}{=}50$ then plateaus; safety remains stable across all settings.}
	\label{tab:ablation_ncal}
\end{table}
